# Enhancing Lung Cancer Treatment Outcome Prediction through Semantic Feature Engineering Using Large Language Models

Mun Hwan Lee, PhD[1], Shaika Chowdhury, PhD [1], Xiaodi Li, PhD[1], Sivaraman Rajaganapathy, PhD [1], Eric W. Klee, PhD[1], Ping Yang, MD, PhD[1], Terence Sio, MD, PhD[1], Liewei Wang, MD, PhD[1], James Cerhan, MD, PhD[1], Nansu N. A. Zong, PhD[1]

[1]Mayo Clinic, Rochester, Minnesota, USA

*Abstract*—Accurate prediction of treatment outcomes in lung cancer is critical yet remains limited by the sparsity, heterogeneity, and contextual overload of real-world electronic health data. Traditional models struggle to capture semantic meaning across multimodal streams, while large-scale fine-tuning approaches are infeasible in clinical workflows. We introduce a novel framework that employs Large Language Models (LLMs) as Goal-oriented Knowledge Curators (GKC) to transform raw laboratory, genomic, and medication data into high-fidelity, task-specific features. Unlike generic embeddings, GKC aligns representation with the prediction objective and integrates seamlessly as an offline preprocessing step, making it practical for deployment in hospital informatics pipelines. Using a lung cancer cohort (N=184), we benchmarked GKC against expert-engineered features, direct text embeddings, and an end-to-end transformer. Our approach achieved a superior mean AUC-ROC of 0.803 (95% CI: 0.799–0.807) and significantly outperformed all baselines. An ablation study further confirmed the synergistic value of combining all three modalities. These results demonstrate that the quality of semantic representation drives predictive accuracy in sparse clinical data settings. By reframing LLMs as knowledge curation engines rather than black-box predictors, our work highlights a scalable, interpretable, and workflow-compatible pathway for advancing AI-driven decision support in oncology.

## I. Introduction

Lung cancer remains the leading cause of cancer-related mortality worldwide, accounting for over 1.8 million deaths annually [1]. Despite advances in diagnosis and therapy, the treatment outcome for lung cancer patients is often poor, with five-year survival rates languishing below 20% in most populations [1, 2]. Accurate prediction of treatment outcome is essential for informed clinical decision-making, guiding treatment selection, risk stratification, and the allocation of scarce medical resources [3, 4].

For decades, the TNM staging system has served as the foundation for lung cancer treatment outcome, providing critical guidance based on the anatomical extent of disease [5, 6]. While it remains the gold standard for initial risk stratification, substantial clinical evidence reveals that patients within the same TNM stage can exhibit markedly different outcomes [7, 8]. This variability arises from complex biological heterogeneity, which includes molecular driver mutations, the tumor immune microenvironment, and key laboratory-based biomarkers [9]. As a result, there is a growing imperative to develop more accurate, personalized models by integrating multi-modal data sources—including laboratory results, genomic profiles, and medication histories—to capture the multifaceted determinants of patient outcomes.

Numerous studies have explored using traditional machine learning to integrate multi-modal structured data, such as concatenating feature vectors from different sources [10-13]. However, these models treat each feature as an isolated, context-free data point. While they can identify correlations, they fail to capture the underlying biological or clinical meaning that expert clinicians extract by synthesizing information holistically across modalities [14]. This gap highlights a critical need for computational paradigms capable of performing deeper semantic integration and interpretive inference on complex biomedical data.

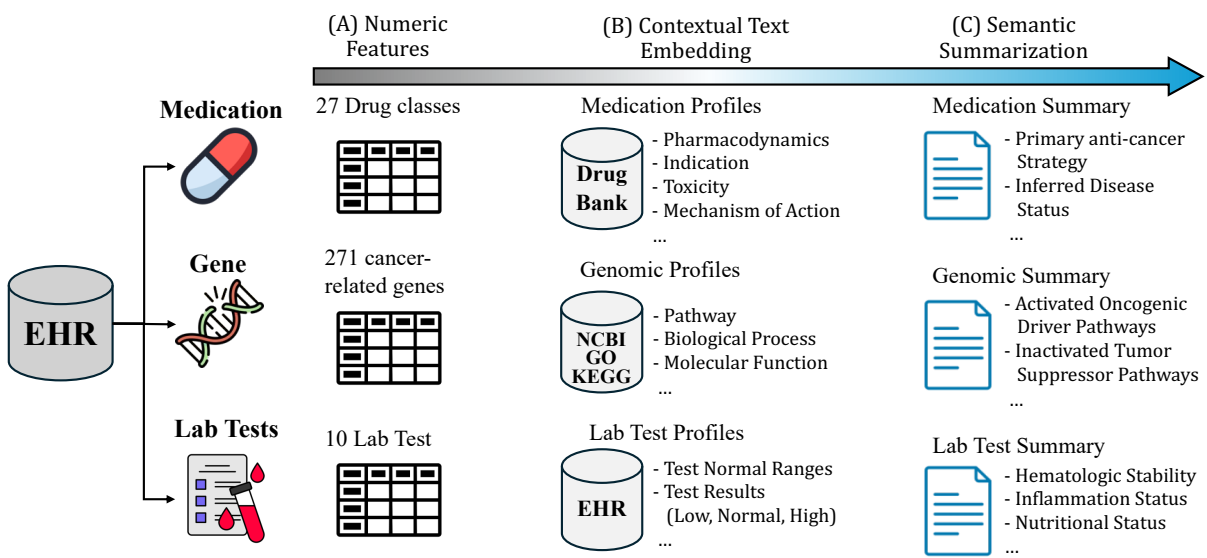

**Figure 1: Overview of the Three Feature Engineering Strategies.** The figure illustrates the three systematic approaches compared in this study. (A) The Expert-Engineered Baseline transforms multi-modal clinical data into a structured numerical feature vector based on pre-defined domain knowledge. (B) The Contextual Text Embedding enriches the raw data with detailed descriptions from biomedical databases to create modality-specific text profiles, which are then embedded. (C) Our Proposed 'Goal-oriented Knowledge Curators' Framework utilizes LLMs to generate a high-fidelity, semantic summary for each modality, and the embeddings of these summaries are used as the final predictive features.

Recent advances in large language models (LLMs) show promise for biomedical applications [15, 16]. Fine-tuned multimodal systems can integrate diverse inputs [17], and advanced prompt engineering techniques [18] can approach specialist performance without task-specific training. However, practical barriers remain. Fine-tuning demands large, high-quality labeled datasets, while sophisticated few-shot prompting depends on many well-crafted exemplars. In clinical prediction, such resources are rare and datasets are typically small, sparse, and heterogeneous. This creates a clear disconnect: the most powerful LLM strategies are often unusable in real-world clinical settings, withholding advanced AI capabilities precisely where they are most urgently needed.

In this study, we address this gap with a framework that uses LLMs for modality-specific semantic summarization. Rather than asking a model to make an end-to-end prediction from all data at once, we first produce goal-aligned summaries per modality (laboratory, genomics, medication). Each summary distills the salient, task-relevant information available in that modality and preserves an interpretable intermediate that can be audited in clinical workflows. The summaries are then embedded and supplied to a lightweight classifier.

Our inputs are assay-grounded, clinician-facing profiles. The genomics modality is a clinically deployed, expert-curated targeted panel of 271 cancer-related genes, reviewed by oncology and molecular genetics teams and focused on actionable driver and resistance biology. In routine care this panel-level genomics is available to a minority of patients due to cost and access constraints, making it pharmacogenomic high-value when present. Serial laboratory tests capture current physiological state (e.g., inflammation and nutritional status).

Structured medication history reflects therapeutic intent and disease trajectory. Modeling these three streams together aligns with how oncology teams actually review patient data.

We evaluate three representation strategies for the same cohort and task (1-year survival): (i) an expert-engineered numerical baseline, (ii) a contextual text-embedding baseline that embeds detailed modality profiles from biomedical knowledge bases, and (iii) the proposed modality-specific summarization approach that first produces goal-aligned modality summaries and then embeds them. All models are assessed with repeated stratified cross-validation under identical protocols.

Our results show a clear hierarchy of representations. Modality-specific summarization achieves the best performance (AUC-ROC 0.803; AUC-PRC 0.859), exceeding both the expert-engineered numerical baseline (AUC-ROC 0.619; AUC-PRC 0.713) and the contextual embedding baseline (AUC-ROC 0.678; AUC-PRC 0.771). An ablation study confirms synergy across modalities, and SHAP analyses indicate balanced contributions from laboratory, genomics, and medication features. Additionally, a long-context BERT baseline with a classification head performed comparably to the contextual embedding baseline, further underscoring that task-aligned summarization, rather than generic self-attention over long profiles, drives the observed gains.

In summary, the key contributions of this work are:

1. We introduce a modality-specific LLM summarization framework that converts assay-grounded clinical inputs

into compact, interpretable features suitable for audit and deployment in real-world oncology.

2. We establish a representation hierarchy, showing that goal-aligned summarization outperforms expert-engineered features and direct embeddings.

3. We demonstrate that preserving pharmacogenomically rich but scarce panel-level genomics alongside laboratory and medication streams improves prediction.

4. We offer empirical evidence including ablations and model comparisons that explicit summarization, not generic self-attention, drives the observed gains in this setting.

## II. METHODS

### A. Study Design and Patient Cohort

This retrospective cohort study was conducted under IRB approval using de-identified electronic health record (EHR) data. We included all adults diagnosed with lung cancer between June 2011 and October 2022 who underwent targeted genomic panel testing. To mitigate lead-time bias, a 90-day landmark was applied; only patients surviving ≥90 days after diagnosis were eligible. Patients were required to have complete laboratory, genomic, and medication data within this 90-day window. After applying these criteria, the final analytic cohort comprised 184 patients.

### B. Cohort Characteristics

The study cohort (N=184) had a mean age of 65.1 years and balanced sex distribution (49% male, 51% female). At the 90-day landmark, 72% had metastatic disease, 46% had major comorbidities, and the median number of mutated genes and prescribed drug classes per patient were 5 and 9, respectively. Overall, 36% of patients died within one year of the landmark. Table 1 summarizes key demographic and clinical characteristics.

### C. Data Sources and Preprocessing

We extracted and integrated three distinct data modalities for each patient within the 90-day window from diagnosis to the landmark time.

- Medication Data (Med): We extracted prescription records for 64 drugs from the EHR. This list was compiled by identifying 38 anti-cancer agents prescribed to our cohort from a list of FDA-approved lung cancer drugs, and adding 26 key supportive care drugs also administered to these patients. This selection was based on a review of the most frequently prescribed medications for this specific patient cohort, reflecting actual clinical practice. To build contextual profiles, we programmatically parsed the full DrugBank database [19] to extract detailed information for each drug, including its description, mechanism-of-action, indication, pharmacodynamics, and toxicity. Individual drugs were then grouped into 27 clinically meaningful classes (e.g., Platinum-Based Chemotherapy, PD-1/PD-L1 Checkpoint Inhibitors, Strong Opioids) based on their mechanism of action.

TABLE I. DEMOGRAPHICS AND CLINICAL CHARACTERISTICS OF THE STUDY COHORT (N=184)

| Characteristic | Value at Landmark |
|---|---|
| **Demographics** | |
| Age, years (Mean ± SD[b]) | 65.1 ± 10.4 |
| Sex, n (%) | |
| Male | 90 (48.9) |
| Female | 94 (51.1) |
| **Clinical Characteristics** | |
| Metastatic Disease, n (%) | 132 (71.7) |
| Key Comorbidities[a], n (%) | 84 (45.7) |
| **Key Modality Summaries** | |
| Laboratory Values | |
| Albumin (g/dL), Mean ± SD | 3.8 ± 0.6 |
| Hemoglobin (g/dL), Mean ± SD | 11.7 ± 2.3 |
| Platelets (K/uL), Mean ± SD | 241.7 ± 119.0 |
| Genomic and Medication Profile | |
| Mutated Genes per Patient, Median [IQR] | 5 [3-7] |
| Drug Classes per Patient, Median [IQR] | 9 [7-11] |
| **Outcome** | |
| Died within 1-Year from Landmark, n (%) | 67 (36.4) |

[a]. Key comorbidities include COPD, Coronary Artery Disease, Congestive Heart Failure, Chronic Kidney Disease, Diabetes, Cerebrovascular Disease, or Connective Tissue Disease diagnosed at or before the 90-day landmark. These clinical variables were used to characterize the cohort.

[b]. SD: Standard Deviation; IQR: Interquartile Range. All data presented as n (%), unless otherwise specified

- Genomic Data (Gene): We utilized data from a targeted sequencing panel of 271 cancer-related genes. This panel is clinically deployed and expert-curated at our institution, covering actionable driver and resistance biology. All gene names were first normalized to their official HUGO Gene Nomenclature Committee (HGNC) symbols. To create rich semantic profiles, we programmatically retrieved annotation data from established biomedical knowledge bases. This included functional summaries from the NCBI Gene database [20], pathway information from the KEGG database [21], and functional annotations (Biological Processes, Molecular Functions) from the Gene Ontology (GO) database [22, 23].

- Laboratory Data (Lab): Based on a literature review in lung cancer, we selected 10 laboratory tests that are well-established [24, 25] as markers for systemic inflammation, nutritional status, and overall physiological resilience. These tests include Albumin, Platelets, and components of the complete blood count with differential (e.g., Neutrophils, Lymphocytes). For each patient, we collected the five most recent values for these tests within the 90-day landmark period. If fewer than five values were available, the series was padded using the most recent available observation (last-observation-carried-forward). Laboratory units were

**(A) Genomic Profile Example:**
1. Mutation Gene: TP53
- Function: This gene encodes a tumor suppressor protein containing transcriptional activation, DNA binding, and oligomerization domains. The encoded protein responds to diverse cellular stresses ...
- KEGG Pathways: MAPK signaling pathway; Cell cycle; p53 signaling pathway; PI3K-Akt signaling pathway; Apoptosis; ...
- Biological Processes: negative regulation of transcription by RNA polymerase II;
- Molecular Functions: transcription cis-regulatory region binding; transcription cis-regulatory region binding;

2. Mutation Gene: KRAS
- Function : This gene, a Kirsten ras oncogene homolog from the mammalian ras gene family, encodes ...
3. Mutation Gene: ...
4. Mutation Gene: ...

**(B) LLM Summarizer Output Example:**
1. Summary: The patient's genomic profile, characterized by concurrent TP53 and KRAS mutations, suggests an aggressive tumor biology and a guarded progress. The activating KRAS mutation will ...
2. Key domains
   2.1 Activated Oncogenic driver pathways: Ras, MAPK, PI3K-Akt , ...
   2.2 Inactivated Tumor suppressor pathways: p53, Cell cycle control, Apoptosis ...
   2.3 Therapeutic implications: Likely resistance to chemotherapy relying on p53 function. ...
3. Key positive factors: None,
4. Key negative factors: (1) Concurrent mutation of a potent oncogene (KRAS) and a critical tumor suppressor (TP53).
   (2) Activation of multiple pro-growth signaling pathways.
   ...

**Figure 2: Comparison of Feature Representation for a Hypothetical Patient's Genomic Profile.** This figure illustrates the distinct inputs for CTE Model and GKC Model using a hypothetical patient profile. (A) The input for CTE Model is a long-form textual profile, created by concatenating detailed descriptions for each dysregulated hallmark domain from public knowledge bases. (B) In contrast, the input for GKC Model is a concise, goal-oriented summary generated by the 'Goal-oriented Knowledge Curators' LLM. This summary synthesizes the raw information from (A) into a high-fidelity narrative focused specifically on implications of treatment outcomes, such as the synergistic effect of the co-mutations and likely immunotherapy resistance. This process transforms noisy, descriptive text into a dense, feature-rich summary for the downstream classifier.

harmonized to institutional reference ranges prior to normalization; missingness, out-of-range entries, and biologically implausible values were screened and resolved by predefined rules.

*D. Feature Engineering Strategies*

We systematically compared three distinct strategies for representing the multi-modal data, each building upon the last. The fundamental difference between our key models is conceptually illustrated using a hypothetical patient's genomic profile in Fig 2.

- ENF (Expert-Engineered Numerical Features) Model: This baseline represents the traditional tabular data-based approach (Fig 1A). Features consisted of a single numerical vector per patient, the five most recent lab values of 10 lab tests, the count of mutated genes, and a binary indicator (0/1) for the prescription of each of the 27 medication classes. Missing medication and genomic entries were treated as absent and encoded as zeros. We normalized the lab values for each lab test to preserve each lab test data's information range because each lab test has a different Unit of test and normal criteria.
- CTE (Contextual Text Embedding) Model: This model was designed to test the value of adding semantic context directly from biomedical databases (Fig. 1B). For each patient, we generated three modality-specific text profiles by concatenating detailed, pre-written descriptions for each relevant data point, as exemplified in Fig. 2A. These profiles were enriched with information from biomedical databases (DrugBank, Gene Ontology, KEGG) for each modality. The complete text profile for each modality was then encoded and concatenated into a final feature vector.
- E2E (End-to-End Transformer) Model: We added a powerful end-to-end baseline. This model uses a state-of-the-art long-context transformer (ModernBERT [29]) fine-tuned directly on the same long-form contextual text profiles used by the CTE model. This approach tests the hypothesis that a large transformer model can automatically learn to summarize and predict without an explicit knowledge curation step.
- GKC (Goal-oriented Knowledge Curators) Model: This is our proposed framework (Fig. 1C). It generates a separate, goal-oriented modality summary for each modality, an example of which is shown in Fig. 2B. The resulting three text summaries were then independently embedded, and their vectors concatenated, creating a multi-channel semantic feature vector that preserves modality-specific insights for the downstream classifier. Compared to CTE, the only architectural difference is the explicit summarization step

*E. LLM-based Semantic Summarization Pipeline*

In our work, 'Semantic Summarization' is defined as a goal-oriented process that involves (1) extracting insights relevant to the prediction target, (2) synthesizing relationships between disparate data points, and (3) generating an interpretive narrative, going beyond simple compression. The core of our proposed model (GKC Model) is a pipeline designed to transform the verbose, context-rich text profiles used by CTE Model into high-fidelity, treatment outcomes related features. As illustrated in Fig. 2, this process uses an LLM to synthesize the raw, descriptive information (Fig. 2A) into a concise, goal-oriented narrative focused specifically on implications for treatment outcomes (Fig. 2B).

This was achieved by prompting the LLM to act as a clinical expert and generate a structured JSON report for each modality. This strategy, a major contribution of this work, was designed to guide the LLM to act as a domain expert for each modality and incorporates three key principles:

- Structured Output: To ensure consistency and enable automated parsing, the LLM was instructed to generate its analysis in a structured JSON format with pre-defined keys. We constrained outputs to a schema with required fields: summary, key domains, therapy implications, positives, and negatives

Identify applicable funding agency here. If none, delete this text box.

- Persona-based Role-playing: The prompt assigned the LLM a specific expert role (e.g., 'clinical pharmacologist' or 'systems biologist') tailored to each modality, aligning its interpretive framing with domain-specific knowledge. This role prompting is used only to steer framing; we do not invoke external tools or knowledge bases at generation time.
- Specialized Task Guidance: A core innovation was specializing the prompts for each data type. For instance, the Gene prompt focused on oncogenic pathway dysregulation, while the Medication prompt assessed treatment intent and disease burden, and the Lab report evaluated inflammation and nutritional status.

This approach forced the LLM to synthesize modality-specific insights through a clinically relevant lens, rather than simply re-stating information. Generation was strictly "from-context": the model summarized only the provided modality profiles. Decoding was deterministic (temperature=0.0; top_k=1) to ensure reproducibility and reduce hallucination variance.

*F. LLMs Model and Embedding Specifications*

We extracted and integrated three distinct data modalities for each patient within the 90-day window from diagnosis to the landmark time.

- LLM for Knowledge Curation: All modality summaries for GKC Model were generated using Gemini 2.0 Flash [26], chosen for its strong performance for Medical expert-annotated hallucination test including chronological ordering, lab data understating and diagnosis prediction [27]. A zero-shot prompting strategy was employed. No retrieval or external knowledge calls were used; summaries are produced solely from the constructed modality profiles.
- Text Embedding and Dimensionality Reduction: All textual data for both CTE Model and GKC Model were converted into dense feature vectors using Google's text-embedding-005 model (Gecko) [28]. The task_type parameter was set to CLASSIFICATION to optimize the embeddings for this downstream task.

*G. Predictive Modeling and Evaluation*

- Classifier: To ensure that our findings were robust and not dependent on a single algorithm, we benchmarked a suite of standard machine learning classifiers. This included tree-based ensembles (XGBoost, Random Forest), a neural network (Fully Connected Neural Networks), a kernel method (Support Vector Machine), and a linear model (Logistic Regression with ElasticNet). The final classifier for reporting all primary results was selected based on the highest and most stable performance in this benchmark, with detailed comparative results presented in Section 4.4.
- E2E Model: We trained a long-context transformer (ModernBERT [29], 8,192-token context) with a classification head under the same splits and evaluation protocol. We fine-tuned ModernBERT-base (~149M parameters) with a classification head using a learning rate of 3e-5, 8 epochs, a linear scheduler with a 10% warm-up ratio, max sequence length 8,192 (ensuring no truncation of our longest profiles), and early stopping on validation AUC-ROC within inner folds. Training used the same repeated stratified CV splits as above.
- Evaluation Protocol: All models were evaluated using 10× repeated 5-fold stratified cross-validation with inner hyperparameter tuning. Primary metrics were Receiver Operating Characteristic Curve (AUC-ROC) and Area Under the Precision-Recall Curve (AUC-PRC), both appropriate for imbalanced outcomes.

*H. Statistical and Interpretation Analysis*

Model performance was evaluated across the 10× repeated 5-fold cross-validation runs. To compare distributions between models, we applied the non-parametric Wilcoxon signed-rank test, which does not assume normality. A p-value < 0.05 was considered statistically significant.

For performance metrics (AUROC and AUPRC), 95% confidence intervals were estimated using bootstrapping with 1,000 resamples, providing robust uncertainty estimates.

To interpret the contribution of each data modality in the best-performing GKC model, we applied SHapley Additive exPlanations (SHAP) [30] This approach quantifies the marginal contribution of features from laboratory, genomic, and medication streams, enabling a transparent assessment of how each modality informs model predictions. SHAP outputs were analyzed both globally (across the entire cohort) and locally (per patient) to support interpretability in clinical contexts.

### III. RESULTS

This section presents the predictive performance of our proposed and baseline models on the task of 1-year survival prediction in lung cancer patients. Our key finding is that the Goal-oriented Knowledge Curator (GKC) model substantially outperformed all baselines, demonstrating that task-aligned semantic summarization provides measurable gains in a data-scarce and heterogeneous clinical setting. Importantly, these improvements address a clinically critical challenge: enabling robust survival prediction even when multi-modal data are incomplete, sparse, or unevenly distributed. We organize the results into three parts: (1) primary comparisons against baseline strategies, (2) robustness of classifier choice, (3) ablation analyses quantifying the synergistic effects of multi-modality, and (4) an in-depth qualitative case study illustrating interpretability and clinical relevance.

*A. Experimental Design and Evaluation Metrics*

All models were evaluated on 1-year survival prediction in a cohort of 184 lung cancer patients, using a 10-times repeated 5-fold stratified cross-validation scheme to provide stable estimates and reduce variance from small sample sizes. Stratification ensured balanced outcome proportions across folds, and all preprocessing steps were contained strictly within training folds to prevent information leakage.

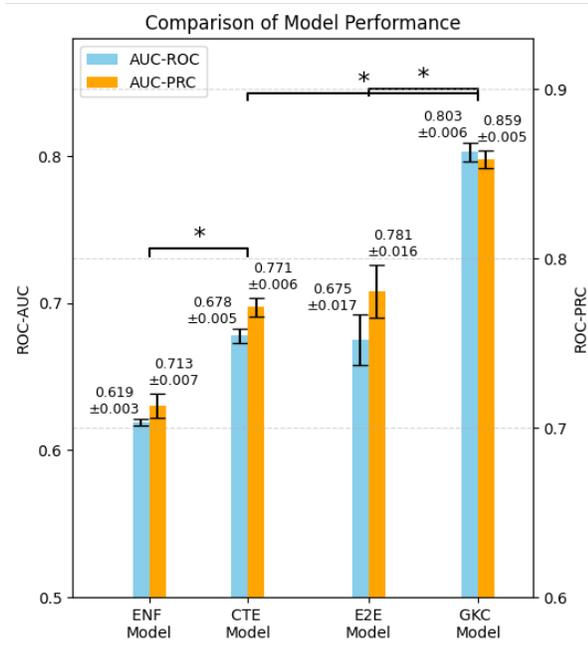

**Figure 3: Comparative performance of feature engineering strategies for 1-year survival prediction in lung cancer.** Box plots show mean AUC-ROC and AUC-PRC across four approaches: numerical encoding (ENF), contextual text embedding (CTE), end-to-end transformer (E2E), and the proposed Goal-oriented Knowledge Curator (GKC). The GKC model significantly outperformed all baselines (p<0.05), achieving both higher discrimination and precision-recall balance. These results highlight that task-aligned semantic summarization provides clinically meaningful improvements over conventional tabular, embedding-based, or end-to-end strategies. Error bars denote standard deviation. Asterisks indicate statistically significant differences.

We report two complementary performance metrics: Area Under the Receiver Operating Characteristic Curve (AUC-ROC), which measures overall discriminative ability, and Area Under the Precision-Recall Curve (AUC-PRC), which is particularly informative under class imbalance. To assess robustness, 95% confidence intervals were obtained via bootstrapping. Statistical significance of model differences was evaluated with the non-parametric Wilcoxon signed-rank test (p < 0.05), ensuring fair and distribution-free comparison across identical validation protocols.

### B. Primary Finding: Superiority of LLM-based Semantic Summarization

Our primary analysis, summarized in Fig. 3, demonstrates a clear and consistent performance hierarchy across the evaluated feature engineering strategies. The expert-engineered numerical baseline (ENF Model) showed only modest discriminative ability, with a mean AUC-ROC of 0.619 (95% CI: 0.618–0.621) and AUC-PRC of 0.713 (95% CI: 0.708–0.718), underscoring the limitations of purely numerical, context-free encoding. Introducing semantic enrichment through contextual biomedical text embedding (CTE Model) improved performance to an AUC-ROC of 0.678 (95% CI: 0.675–0.681) and AUC-PRC of 0.771 (95% CI: 0.768–0.775), highlighting the value of leveraging biomedical knowledge bases to capture context that numeric features alone cannot provide.

To test whether a large-scale end-to-end approach could overcome these limitations, we trained a long-context transformer (E2E Model, ModernBERT, 8k tokens) directly on the same contextual inputs used in the CTE model. Despite its architectural capacity, the model achieved similar performance (AUC-ROC 0.675, 95% CI: 0.663–0.685; AUC-PRC 0.781, 95% CI: 0.772–0.791), with no statistically significant difference from CTE (Wilcoxon p=0.770 for AUC-ROC; p=0.105 for AUC-PRC). This finding indicates that simply scaling up the architecture without explicit knowledge curation does not overcome the inherent challenges of sparse, heterogeneous clinical data.

By contrast, our proposed Goal-oriented Knowledge Curator (GKC) model achieved a decisive leap in predictive accuracy, reaching an AUC-ROC of 0.803 (95% CI: 0.799–0.807) and AUC-PRC of 0.859 (95% CI: 0.855–0.862). Improvements were statistically significant compared with all baselines (p<0.05 across both metrics), demonstrating that explicit, task-aligned summarization captures clinically salient information that is missed by both numerical aggregation and generic embeddings. To examine whether this advantage was merely a function of LLM choice, we tested a biomedical-specific variant of GKC using MedFound-176B [31]. This configuration achieved an AUC-ROC of 0.752 and AUC-PRC of 0.832, higher than traditional baselines but still inferior to our optimal GKC setup. Together, these findings indicate that the core strength of our framework lies not in model scale or domain specialization, but in its design principles of goal-aligned summarization and efficient offline representation.

From a clinical perspective, these findings underscore that survival prediction in lung cancer depends not merely on the quantity of data available but on how heterogeneous inputs are distilled into meaningful, interpretable representations. The GKC framework reflects the way oncologists synthesize laboratory results, genomic profiles, and medication histories into concise, decision-ready narratives. In this way, our results demonstrate that the quality of representation, rather than the raw volume of data, is the critical determinant of predictive accuracy in this setting.

### C. Robustness of the Predictive Classifier

To evaluate the optimal classifier for LLM-derived semantic features, we benchmarked multiple standard machine learning algorithms on the identical feature set produced by the GKC model. As shown in Fig. 4, tree-based ensemble methods provided the most reliable performance. XGBoost achieved the highest mean AUC-ROC of 0.800 (±0.006) and a corresponding AUC-PRC of 0.859, followed by Random Forest with an AUC-ROC of 0.760 (±0.007). In contrast, high-capacity models such as Fully Connected Neural Networks yielded both lower accuracy (AUC-ROC 0.720 ±0.012) and greater variance across folds, reflecting overfitting in this small-sample regime (N=184). These findings demonstrate that tree-based ensembles are particularly well suited to dense embedding features in sparse clinical datasets, where their built-in regularization and capacity for nonlinear interaction modeling provide stability without sacrificing discriminative power. Based on this evidence, XGBoost was chosen as the primary classifier for all subsequent analyses. Clinically, this result underscores that methodological rigor in model choice can substantially impact predictive reliability, particularly in rare or data-limited patient cohorts.

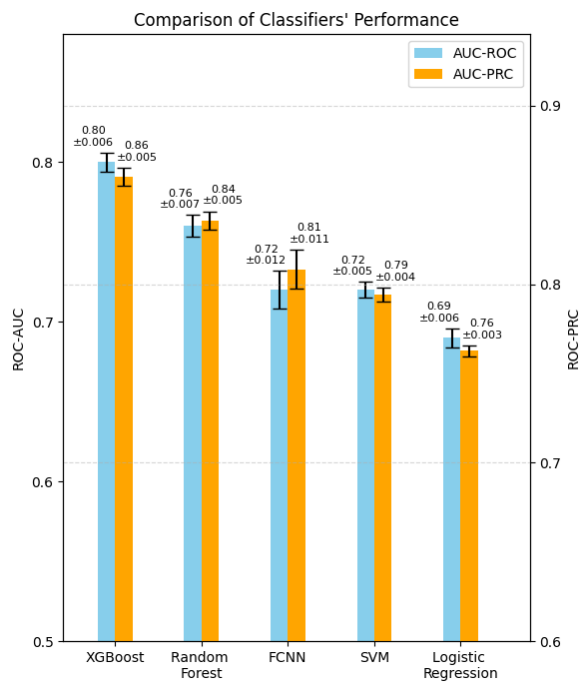

**Figure 4: Comparative Performance of Predictive Classifiers on GKC Model Features.** Mean AUC-ROC and AUC-PRC for five standard machine learning algorithms trained on the LLM-generated features from GKC Model. XGBoost achieved the highest performance across both metrics, justifying its selection as the primary classifier for this study.

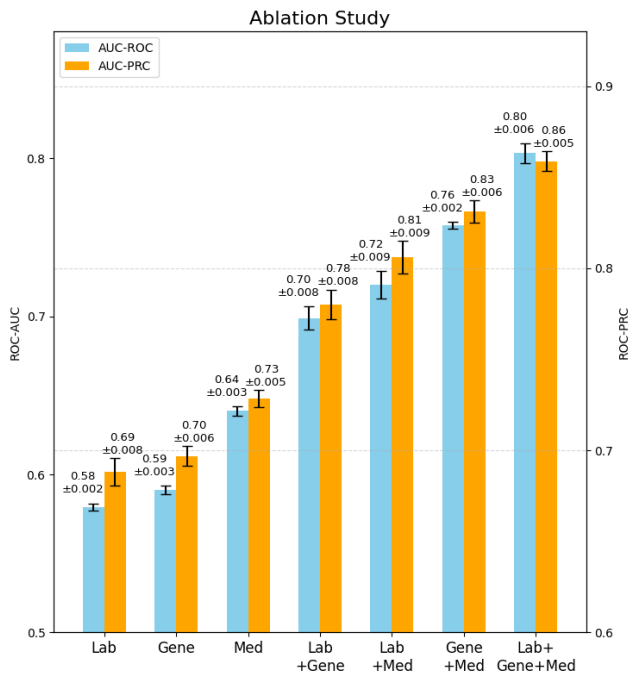

**Figure 5: Ablation Study of Modality Contribution to Treatment Outcome Prediction Performance.** Mean AUC-ROC and AUC-PRC performance from an ablation study on GKC Model, evaluating all combinations of Lab, Gene, and Med data. The results show a clear stepwise improvement as modalities are combined, with the full three-modality model achieving the highest performance. This demonstrates that each modality provides complementary, non-redundant information.

*D. Synergistic Effect of Multi-Modality (Ablation Study)*

We next conducted an ablation study to quantify the individual and combined contributions of laboratory, genomic, and medication modalities. Fig. 5 summarizes results across all seven possible modality combinations. Among single modalities, medication history emerged as the strongest individual predictor (AUC-ROC 0.640; AUC-PRC 0.742), outperforming both genomic data (AUC-ROC 0.590; AUC-PRC 0.701) and laboratory tests (AUC-ROC 0.580; AUC-PRC 0.695). This suggests that therapeutic intent and treatment trajectory encode critical early survival signals.

The most striking finding was the synergistic benefit of combining modalities. Performance improved monotonically as more data streams were integrated, with the Gene+Med pairing yielding a marked boost (AUC-ROC 0.760; AUC-PRC 0.830). Ultimately, the full tri-modal integration (Lab+Gene+Med) delivered the best results, achieving AUC-ROC 0.800 and AUC-PRC 0.860. These gains were statistically significant (Wilcoxon $p<0.05$ across comparisons), confirming that each modality provides unique, complementary information. Clinically, this finding validates the multidisciplinary perspective already employed by oncology teams, where survival is best understood by synthesizing pharmacologic, genomic, and physiologic evidence rather than treating them in isolation.

*E. Modality Importance in the Final Model (SHAP Analysis)*

To better understand why the tri-modal GKC model succeeds, we decomposed the predictive signal using SHAP. Results are shown in Fig. 6. At the global level, all three modalities contributed almost equally (Medication 33.2%, Gene 33.9%, Lab 32.9%), with pairwise tests confirming that none dominated (all $p>0.05$). This balance demonstrates that the model does not rely on a single "silver bullet" input, but instead integrates complementary signals across domains.

At the patient level, the boxplot distribution (Fig. 6B) revealed natural variability, where some patients' predictions were driven more by medications and others by genomics or labs. This reflects real-world oncology practice, in which certain patients' outcomes hinge more on pharmacologic trajectories (e.g., checkpoint inhibitors), while others are better explained by molecular drivers or physiologic frailty. Clinically, this balanced reliance is reassuring: it indicates that the GKC framework learns to weigh whichever modality is most informative for each individual, much like a multidisciplinary care team does in practice. This modality-level interpretability provides both mechanistic validation and practical transparency for potential deployment.

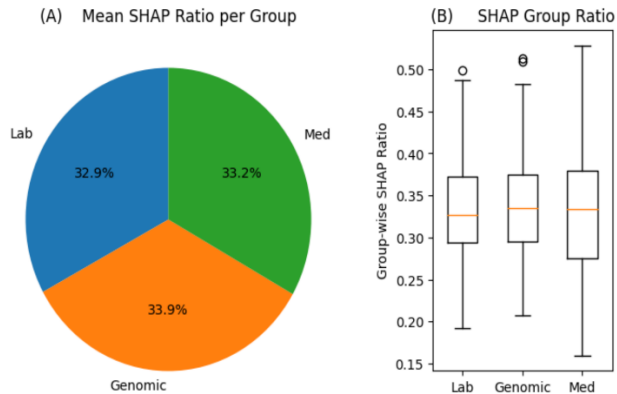

**Figure 6: Balanced Contribution of Multi-Modal Features to Treatment Outcome Prediction.** SHAP analysis of the final prediction model. (A) The mean global feature importance is almost equally distributed among the Lab, Gene, and Medication modalities. (B) Box plots showing the distribution of per-patient contributions confirm this balanced interplay. This result provides a mechanistic explanation for our model's success, demonstrating that it learns by flexibly integrating all three independent information streams rather than relying on a single dominant modality.

*F. In-depth Qualitative Case Study*

To illustrate the mechanism of "knowledge curation" beyond numerical performance, we present a representative patient case with seven oncogenic alterations (REL, RICTOR, MDM2, CDK4, ALK, ATR, KRAS). The raw input consisted of a concatenated, verbose profile from biomedical databases, spanning several thousand tokens and filled with redundant or tangential associations.

The GKC pipeline transformed this into a concise, structured JSON report as shown in Fig. 7, performing three expert-like interpretive tasks that highlight the framework's potential for clinically meaningful synthesis:

1. Noise filtration: irrelevant entries (e.g., non-cancer functions of REL) were removed, and redundant pathway mentions consolidated.

2. Clinical inference potential: the LLM highlighted the co-occurrence of KRAS mutation and MDM2 amplification as a synergistic "double-hit," suggesting its potential negative prognostic implication. This relationship is not explicitly present in the raw text and indicates that the system can surface higher-order associations resembling what is often emphasized in expert tumor board discussions, rather than being limited to literal text extraction.

3. Holistic synthesis: the seven mutations were contextualized into an "aggressive, therapy-resistant phenotype with genomic instability," distilling disparate evidence into a coherent prognostic picture.

This restructuring turned an unwieldy gene-centric list into a goal-aligned prognostic summary, with fields such as prognostic summary and key negative factors. Importantly, the transformation was deterministic and reproducible, minimizing stochastic variability often associated with LLMs.

---

**Gene Report Example**

```
{
1. "prognostic_summary": "This patient's genomic profile suggests an aggressive tumor phenotype with a potentially poor prognosis. The presence of mutant KRAS alongside MDM2 amplification indicates strong activation of proliferative and survival signaling pathways, especially MAPK, PI3K-Akt, and p53, which is further augmented by CDK4 activation. Concurrent REL activation suggests increased inflammatory response and NF-kB signaling, promoting tumor survival and immune evasion. ATR mutations may compromise DNA damage repair, promoting genomic instability and potentially accelerating tumor progression, while RICTOR activation may further contribute to cell growth and survival. The interplay of these mutations suggests a highly proliferative, therapy-resistant tumor, which might be less responsive to conventional treatments.",
2. "key_prognostic_domains": {
2.1 "oncogenic_driver_pathways_activated": [
    "Ras signaling pathway",
    "PI3K-Akt signaling pathway",
    "Cell cycle (CDK4)",
    "NF-kB signaling pathway (REL)",
    "mTORC2 signaling (RICTOR)"   ],
2.2 "tumor_suppressor_pathways_inactivated": [
    "p53 signaling pathway (MDM2 mediated)",
    "DNA damage repair (ATR)"    ],
2.3 "therapeutic_implications": [
    "Potential resistance to therapies targeting p53",
    "Potential sensitivity to MEK or PI3K inhibitors (KRAS)",
    "Potential sensitivity to CDK4/6 inhibitors (CDK4)",
    "Potential sensitivity to ATR inhibitors but resistance if resistance mutations develop." ]
  },
3. "key_positive_factors": [],
4. "key_negative_factors": [
    "Concurrent KRAS mutation and MDM2 amplification.",
    "ATR mutation suggests genomic instability.",
    "REL activation promoting survival and immune evasion." ]}
```

**Figure 7: Gene Report Example generated by the Goal-oriented Knowledge Curator (GKC) framework.** The structured output summarizes seven oncogenic alterations into a concise prognostic profile, highlighting activated driver pathways, inactivated tumor-suppressor functions, and therapeutic implications. Key negative factors such as concurrent KRAS mutation with MDM2 amplification are explicitly identified, demonstrating how GKC transforms verbose genomic data into clinically interpretable insights.

Clinically, this case illustrates how the GKC framework can approximate aspects of oncologists' interpretive practices: filtering noise, identifying potential synergies, and restructuring raw molecular data into actionable, interpretable insights. This qualitative evidence complements the quantitative performance lift, showing that our approach achieves its impact not by brute computational scale but by aligning LLM outputs with clinical interpretive practices.

## IV. DISCUSSION

This study demonstrates that task-aligned semantic summarization by large language models (LLMs) can unlock predictive signal from sparse, heterogeneous real-world oncology data. By positioning LLMs as "Goal-oriented

Knowledge Curators," we showed consistent and significant improvements over numerical baselines, contextual embeddings, and even a powerful long-context transformer baseline. These findings establish representation quality—not just data quantity—as a critical driver of predictive accuracy in clinical AI.

A central insight of this work is that the "small data" problem extends beyond limited sample size. Conventional numerical features strip away latent clinical meaning, whereas our modality-specific summaries preserve the same interpretive context that oncologists extract when integrating evidence across labs, genomics, and medications. The superior performance of our framework demonstrates that models can approximate expert-like synthesis when provided with concise, task-aligned representations.

Our ablation analysis reinforces the importance of holistic integration. Each modality contributed distinct perspectives: medications captured therapeutic trajectory, genomics provided the biological blueprint, and laboratories reflected the dynamic physiological state. The highest accuracy was achieved only when all three were combined (AUC-ROC 0.803, AUC-PRC 0.859). Importantly, the synergy observed between genomics and medications highlights how biological and therapeutic signals must be interpreted together to approximate clinical decision-making.

A noteworthy and somewhat counterintuitive finding was that our general-purpose LLM summarizer outperformed a biomedical-specific model (MedFound-176B). We interpret this to reflect the nature of our task: the challenge is not simply biomedical fact retrieval, but complex instruction-following and synthesis into prognostic narratives. General-purpose LLMs, trained on diverse textual forms and interpretive patterns, may be particularly suited for such task-oriented knowledge curation. This finding suggests that model scale and domain specialization are secondary to framework design—goal-aligned summarization and efficient offline representation are the real drivers of performance.

Mechanistically, SHAP analysis revealed a balanced contribution across all modalities. Rather than depending on a single dominant input, the model flexibly combined independent, high-fidelity streams. This validates our design choice to preserve modality-specific structure before integration, ensuring that clinically meaningful interactions are not prematurely lost.

Despite these encouraging results, several limitations must be acknowledged. First, this proof-of-concept study was limited to a single institution with a modest sample size (N=184). While this setting magnified the value of semantic summarization in overcoming data scarcity, external validation on larger, multi-institutional cohorts is essential. Second, our inputs were restricted to structured and semi-structured data; unstructured clinical narratives and imaging, which contain valuable information such as staging details and radiologic response, remain untapped. Third, the current implementation relies on proprietary APIs. While we demonstrated that the operational cost (~$0.001 per patient) and latency ($\approx 1.6$ seconds per patient) are negligible for deployment, further work with open-source models will be critical to ensure long-term reproducibility and accessibility.

Future directions include validation across cancers and institutions, integration of additional modalities such as pathology and imaging, and exploration of deployment pathways into clinical decision-support workflows. The modularity of our pipeline makes it inherently adaptable: as open-source LLMs and embedding models improve, they can be swapped into our framework without altering its core principles. Ultimately, by aligning representation with clinical synthesis and abstraction, our approach offers a scalable foundation for precision oncology in real-world practice.

## V. Conclusion

In this work, we confronted the challenge of prediction of treatment outcomes from sparse, real-world clinical data. We hypothesized that the limitations of traditional numerical features could be overcome by leveraging LLMs to generate high-quality, semantically rich features. To test this, we developed a novel 'Goal-oriented Knowledge Curator' framework, where LLMs transform raw, multi-modal data (Lab, Genomics, Medication) into modality-specific modality summaries. Our experiments systematically demonstrated the superiority of this approach, with our final model achieving best performance with AUC-ROC of 0.803 and AUC-PRC of 0.859. The primary contribution of this study is the finding that the optimal role for LLMs in this domain is as goal-oriented knowledge curators, generating high-fidelity intermediate features for a downstream classifier. By capturing the complementary interplay among the biological 'blueprint' (Genomics), physiological 'state' (Lab), and therapeutic 'trajectory' (Medication), our approach delivers a new computational paradigm with immediate clinical relevance and broad future applicability for precision medicine in data-scarce environments.

ACKNOWLEDGMENT